\documentclass[letterpaper, 10 pt, journal, twoside]{ieeetran}

\IEEEoverridecommandlockouts                              

\usepackage{graphics} 
\usepackage{epsfig} 
\usepackage{times} 
\usepackage{amssymb}  
\usepackage{bm}
\usepackage{xcolor}
\usepackage{cite}
\usepackage{pifont}

\usepackage{ulem} 
\usepackage{hyperref}
\hypersetup{pdfstartview=FitH,
            colorlinks=true,
            linkcolor=red,
            anchorcolor=blue,
            citecolor=black
            }

\usepackage{amsthm}
\usepackage{float}
\usepackage{booktabs}
\usepackage{multirow}
\usepackage{makecell}
\usepackage{enumerate}
\usepackage{leftindex}
\usepackage{mathtools}
\usepackage{mathrsfs}
\usepackage[utf8]{inputenc}
\usepackage[caption=false,font=footnotesize]{subfig}
\usepackage{graphicx}
\usepackage{pifont}
\usepackage{soul}
\usepackage[ruled,linesnumbered]{algorithm2e}  
\usepackage{bbding}

\usepackage{graphicx}
\newcounter{RNum}
\renewcommand{\theRNum}{\arabic{RNum}}
\usepackage{tabularray}
\usepackage{makecell}

\theoremstyle{plain}

\theoremstyle{definition}

\newcommand{\Remark}{\noindent\textbf{Remark}~\refstepcounter{RNum}\textbf{\theRNum}: }
\newcommand{\NoOne}[1]{\textcolor{red}{#1}}
\newcommand{\NoTwo}[1]{\textcolor{green}{#1}}
\newcommand{\NoThree}[1]{\textcolor{blue}{#1}}
\makeatletter 
\pretocmd\@bibitem{\color{black}\csname keycolor#1\endcsname}{}{\fail}
\newcommand\citecolor[1]{\@namedef{keycolor#1}{\color{blue}}}
\makeatother
\newcommand{\re}{\mathbb{R}}

\soulregister\NoOne7
\soulregister\NoTwo7
\soulregister\NoThree7
\soulregister\Remark7

\title{
FRTree Planner: Robot Navigation in Cluttered and Unknown Environments with Tree of Free Regions
}

\author{Yulin Li$^{1,*}$, Zhicheng Song$^{2,*}$, Chunxin Zheng$^{2}$, Zhihai Bi$^{2}$, Kai Chen$^{2}$,\\ Michael Yu Wang$^{3}$, \textit{Fellow, IEEE,} and Jun Ma$^{1}$ 
 \thanks{$^{*}$indicates equal contribution.}%
\thanks{$^{1}$Yulin Li and Jun Ma are with the Division of Emerging Interdisciplinary Areas, The Hong Kong University of Science and Technology, Hong Kong SAR, China (e-mail: yline@connect.ust.hk; jun.ma@ust.hk)
}%
\thanks{$^{2}$Zhicheng Song, Chunxin Zheng, Zhihai Bi, and Kai Chen are with the Robotics and Autonomous Systems Thrust, The Hong Kong University of Science and Technology (Guangzhou), Guangzhou, China (e-mail: zsong469@connect.hkust-gz.edu.cn; czheng739@connect.hkust-gz.edu.cn; zbi217@connect.hkust-gz.edu.cn; kchen916@connect.hkust-gz.edu.cn)}
\thanks{$^{3}$Michael Yu Wang is with the School of Engineering, Great Bay University, China (e-mail: mywang@gbu.edu.cn)}
}

\begin{document}

\maketitle
\thispagestyle{empty}
\pagestyle{empty}


\begin{abstract}
In this work, we present FRTree planner, a novel robot navigation framework that leverages a tree structure of free regions, specifically designed for navigation in cluttered and unknown environments with narrow passages. 
The framework continuously incorporates real-time perceptive information to identify distinct navigation options and dynamically expands the tree toward explorable and traversable directions. This dynamically constructed tree incrementally encodes the geometric and topological information of the collision-free space, enabling efficient selection of the intermediate goals, navigating around dead-end situations, and avoidance of dynamic obstacles without a prior map.
Crucially, our method performs a comprehensive analysis of the geometric relationship between free regions and the robot during online replanning. In particular, the planner assesses the accessibility of candidate passages based on the robot's geometries, facilitating the effective selection of the most viable intermediate goals through accessible narrow passages while minimizing unnecessary detours. By combining the free region information with a bi-level trajectory optimization tailored for robots with specific geometries, our approach generates robust and adaptable obstacle avoidance strategies in confined spaces.
Through extensive simulations and real-world experiments, FRTree demonstrates its superiority over benchmark methods in generating safe, efficient motion plans through highly cluttered and unknown terrains with narrow gaps. The open-source project can be found at \url{https://github.com/lyl00/navigation_with_tree_of_free_regions}.

\end{abstract}

\section{Introduction} \label{sec:intro}
Over the past decades, the field of motion planning in robotics has witnessed significant advancements, which markedly improve the mobility and flexibility of mobile robots when navigating through complex environments \cite{8758904,faster,10363679, 10607111}. Despite these achievements, the development of an autonomous navigation system capable of operating efficiently in cluttered and fully unknown environments remains a formidable challenge in research. In such settings, the robot primarily encounters two types of challenges. 
First, when navigating through narrow passages, the navigation system must assess if a narrow gap is traversable based on the robot's geometry, and then generate safe and effective motion plans to pass through it. Otherwise, it may lead to overly conservative maneuvers to bypass accessible narrow passages, or failure to find a feasible path. 
Second, in fully unknown environments with limited sensor range, the system must autonomously make informed decisions based on local perception information at each replanning phase. 
This involves selecting intermediate goals, overcoming dead-end situations, and avoiding unforeseen dynamic obstacles, which render it even more challenging in cluttered environments.



In our previous work \cite{li2024collisionfreetrajectoryoptimizationcluttered}, a bi-level trajectory optimization algorithm is proposed to generate collision-free trajectory by constraining robots with specific geometries to be contained within the free space over the entire optimization horizon.
However, this method relies on pre-decomposition in a known environment, which involves sampling points in the free space for extracting free regions until sufficient regions are generated to approximate the entire free space.
In cluttered environments with complex obstacle layouts, an excessive number of regions are required to represent the free space. This results in a dense graph with redundant information that complicates the search for a reference path and also increases the computational burden. Additionally, effectively sampling points at narrow gaps poses significant challenge, which subsequently leads to formation of low-quality and non-traversable regions in narrow passage. Essentially, it is the typical case that these challenges render the optimization problem rather difficult to solve, and even lead to failure of convergence in some scenarios.
Furthermore, since this method lacks mechanisms to update the graph using local perception information and explore the optimal directions, its deployment in certain circumstances is inherently restricted, for example, when the robot gets trapped in confined spaces or navigates around moving obstacles in high-dynamic environments.

\begin{figure*}[t]	
	\centering
    \vspace{3pt}
	\includegraphics{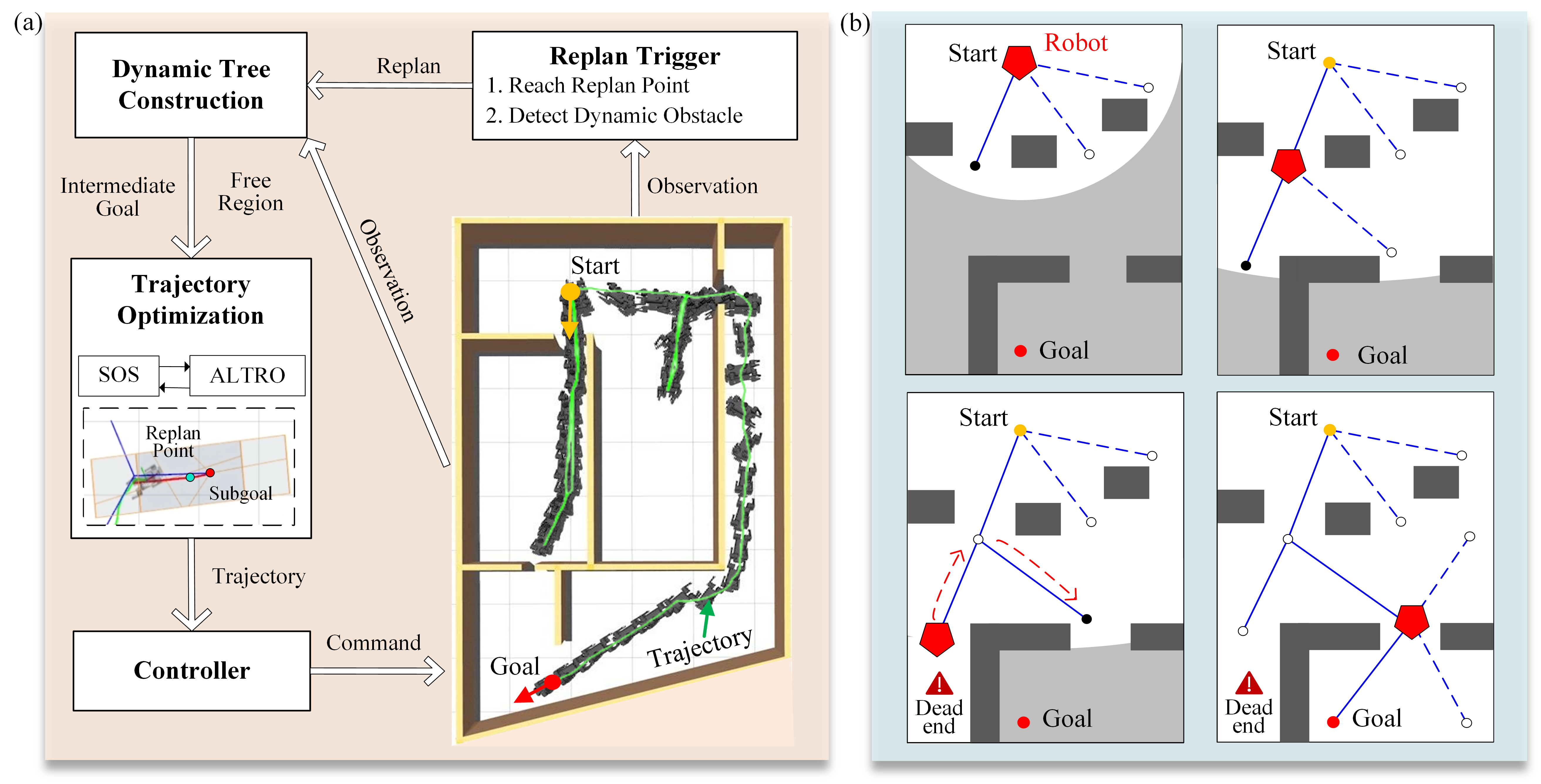}
        \vspace{10pt}
	\setlength{\abovecaptionskip}{-10pt} 
	\caption
{
Overview of the proposed navigation framework. (a)  Illustration of the framework pipeline. At each replanning phase, a tree of free regions is dynamically constructed to efficiently embed information about the free space and potential exploration directions. The next feasible and explorable intermediate goal is inferred and fed into the subsequent geometry-aware bi-level trajectory optimization framework to achieve safe and efficient navigation in unknown and cluttered environments with narrow passages and bug traps. (b) Visualization of the navigation process with limited sensor range. As navigation progresses, the free region tree is continuously updated that records visited and dead-end areas. This enables the consistent selection of suitable intermediate goals, ensuring safe and efficient navigation to the destination.}
\label{fig:framework}
\end{figure*}

To address all the aforementioned limitations, this paper extends our previous work~\cite{li2024collisionfreetrajectoryoptimizationcluttered} and proposes FRTree planner for robot navigation in cluttered and unknown environments.
The overview of the proposed navigation framework is shown in Fig. \ref{fig:framework}, and the main contributions of our work are as follows: 
\begin{itemize}
\item We propose a novel map-free robot navigation framework that effectively exploits the topology of free space by constructing a tree of free regions. This approach facilitates online replanning of safe and efficient goal-directed trajectories in unknown and cluttered environments with limited sensor range.

\item Real-time perception information is continuously integrated to expand the tree toward directions that can be explored and transversed, and this allows the robot to effectively select the most viable intermediate goals, mitigate dead-end situations, and avoid dynamic obstacles without relying on a prior map. 

\item The framework efficiently identifies narrow passages that are traversable, tailored to the robot's specific geometry. Free regions generated along directions leading to these narrow passages are integrated with a backend geometry-aware, collision-free trajectory optimization. This integration allows for more robust and effective generations of adaptable obstacle avoidance behaviors in highly cluttered spaces.
\end{itemize}

\section{RELATED WORKS}
Sampling-based methods have long been favored in motion planning for their efficiency in finding paths in cluttered environments. Techniques like rapidly exploring random tree (RRT) and probabilistic roadmap (PRM) create paths by connecting sampled points in the collision-free space towards the goal. Although these methods offer probabilistic completeness, they lack asymptotic
optimality. To address this, extensions such as RRT*\cite{rrtstar} and RRT*-Smart\cite{rrt_smart} have been developed to guarantee optimal solutions provided with enough samples. However, the sampling-based approach requires substantial computational resources, making it impractical for real-time recomputation in dynamic environments. To address this, variants of these methods regenerate feasible paths as conditions change \cite{local_rrt,rrtx}. They typically modify the initial path by continuously refining the search tree when new information comes. Still, a large search tree needs to be maintained and updated at each replanning phase. 

Recent improvements in computational power and optimization algorithms have highlighted the potential of optimization-based methods for generating safe and effective trajectories in complex environments\cite{9765821,9490372,9353198}.
In cluttered environments with complex obstacle layouts, the spatial decomposition of the free space is widely explored \cite{10417140,9560773,9718137}.
Various techniques exist to obtain such decompositions to efficiently approximate the free space \cite{7839930,toumieh2022voxel}, enabling effective enforcement of collision avoidance constraints in the subsequent calculation of safe trajectories. In \cite{7839930}, sampled points on the polynomial trajectory are enforced to be within the extracted free region sequence, which is embedded into a minimum snap
framework, adding a safety layer to the optimized motion. Extensions of this idea use polyhedral outer representations to express the whole trajectory utilizing the convex hull property. Specifically, control
points of Bernstein basis \cite{faster}, B-spline basis\cite{8758904}, and MINVO basis\cite{9490372} 
can be confined within the safe corridor to ensure the entire trajectory’s
safety. Yet, these methods only generate free regions along the reference path searched from dilated obstacle information without considering specific robot geometries \cite{9999335,7839930,10599811}, typically leading to failure to find feasible paths, especially in cluttered environments with narrow passages.
To address this, a bi-level trajectory optimization framework has been proposed to consider the robot's specific geometry, which ensures that the robot remains within the sequence of free regions throughout the optimization process \cite{li2024collisionfreetrajectoryoptimizationcluttered}.

In fully unknown settings, maintaining an incrementally fused map, such as an occupancy map or ESDF map, is necessary due to limited sensor range and perception noise. The local reference is regenerated based on the updated map fused with new sensor data, which is then refined into feasible, collision-free trajectories via backend optimization. Although effective and widely applied in many advanced navigation systems \cite{8276241,10599811,faster}, this two-step navigation framework can suffer from additional computational burden due to preprocessing and accumulated mapping errors.
Alternatively, a graph of connected free regions has the potential to efficiently explore the spatial structure and represent larger portions of the free space compared to discrete grids \cite{9561460, science-gcs, li2024collisionfreetrajectoryoptimizationcluttered}. However, existing methods either rely on a pre-generated graph of free regions and lack mechanisms to update the graph locally, or their graphs represent each node as a single region generated from a sample point. These approaches do not fully utilize the topology information of the free space to examine traversability and guide exploration, leading to difficulties in finding feasible paths and resulting in conservative maneuvers in narrow and cluttered spaces.

\begin{table}[t]
    \centering
    \caption{NOMENCLATURE}
\begin{tabular}{cc}
    \toprule
     \textbf{Symbols}& \textbf{Descriptions} \\
     \midrule
     $\boldsymbol{\mathcal{P}}_{\boldsymbol{p}}$ & Point cloud data perceived at position $\boldsymbol{p}$\\
     $\boldsymbol{\mathcal{F}}_{\boldsymbol{p}}$ & Set of all feature points extracted at position 
     $\boldsymbol{p}$\\
     $\boldsymbol{\mathcal{Q}}_{n_i}$& Sequence of free regions stored in node $i$\\
     $\boldsymbol{\mathcal{S}}_{n_i}$ & Set of all child nodes of node $i$\\
     $\boldsymbol{\mathcal{R}}_{n_i}$ & Set of interesting direction grown from node $i$ \\
     $\boldsymbol{\mathcal{V}}$ & Set of all free regions marked for \texttt{visited} nodes\\
     $\boldsymbol{\mathcal{D}}$ & Set of all free regions marked for \texttt{dead} nodes\\
     $\boldsymbol{C}(\mathcal{Q}) \in \re^{3}$ & Geometric center of a polytopic free region $\mathcal{Q}$\\
     $\boldsymbol{p}_{n_i}^r \in \re^{3}$ & Replan point for node $i$\\
     $\boldsymbol{p}_s \in \re^{3}$ & Start position \\
     $\boldsymbol{p}_g \in \re^{3}$ & Goal position \\
     \bottomrule
\end{tabular}
    \label{tab:some_definition}
\end{table}

\section{Methodology}
This work aims to develop a navigation framework that generates collision-free motion plans for a robot with specific geometry to achieve real-time goal-directed maneuvers in cluttered and fully unknown environments. The proposed navigation system is capable of continuously integrating new sensory information within the constraints of limited sensor range, to replan safe and efficient trajectories toward the intended goal configuration. This entails the system's proficiency in efficient planning and exploration, selection of the most viable path, generation of safe and dynamically feasible trajectories in narrow spaces, recovery from trapping into local optima, and adaptation to unexpected or changing environments.
Specifically, the framework introduces an online replanning mechanism that examines the geometric layout of the free space, as illustrated in Fig. \ref{fig:framework}, which primarily relies on the iterative execution of three sequential steps at each replanning phase:
\begin{itemize}
    \item Dynamical construction of the tree of free regions $\mathcal{T}$.
    \item Intermediate goal selection with updated $\mathcal{T}$.
    \item Geometry-aware collision-free trajectory optimization.
\end{itemize}
Pertinent notations used in this work are listed in Table \ref{tab:some_definition} for convenience. Starting from $\boldsymbol{p}_s$ in an initially unknown environment with dense obstacles, we progressively gather more information about the surroundings as the robot navigates the environment. This is achieved by incrementally constructing a tree of free regions  $\mathcal{T}$ rooted at $\boldsymbol{p}_s$. Utilizing this updated graph, we continuously choose appropriate intermediate goals and optimize the local trajectory, and this enables the robot to safely and efficiently reach $\boldsymbol{p}_g$ in complex environments.


\subsection{Dynamic Tree Construction}\label{sec:dynamic_tree}
In this section, we explain the tree structure and how to update it online using real-time perception data. In our proposed framework, we represent a node as a potential direction for exploration from its parent node, together with the geometric information of the free space extracted along that direction. Each node encapsulates a specific path and its associated free regions, enabling an efficient representation of free space while minimizing information overhead compared to conventional occupancy grid maps. Specifically, for each node $n_i\in\mathcal{N}$, we extract a sequence of free regions $\boldsymbol{\mathcal{Q}}_i$ that represents the spatial structure of the exploration direction. A replan point $\boldsymbol{p}_{n_i}^r$ is then associated with $n_i$ as the intermediate goal for trajectory optimization when the node is selected for further exploration. The process of dynamically constructing a tree with such nodes is illustrated in Algorithm \ref{alg:cfto}. 

\begin{figure}[t]	
\centering
    \vspace{3pt}
\includegraphics[trim=0.1cm 0.1cm 0 0.1cm, clip,width=1\linewidth]{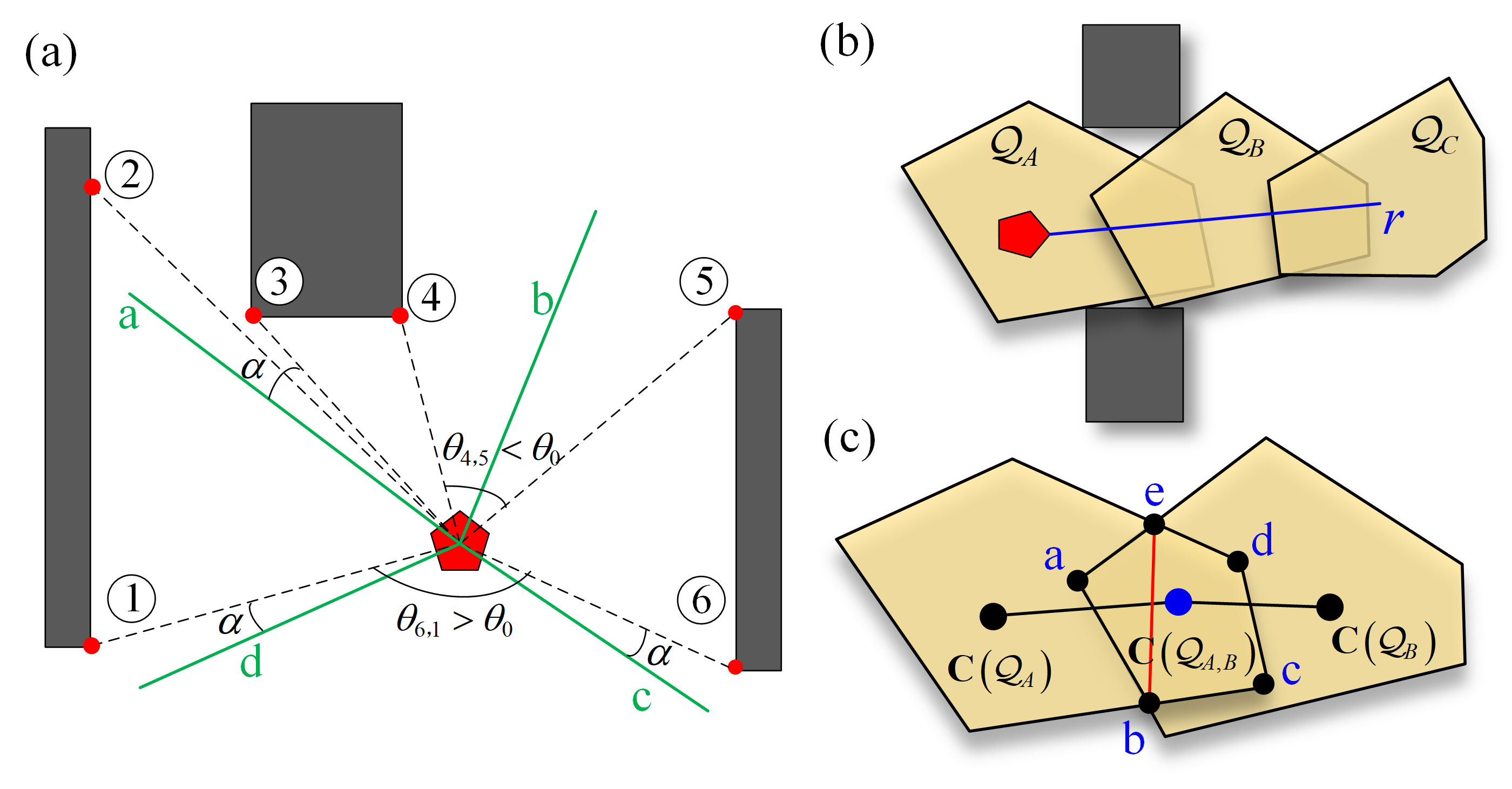}
\setlength{\abovecaptionskip}{-0pt} 
\caption
{Illustration of the dynamic tree construction. (a) Visualization of the process for identifying the interesting directions.
(b) Depiction of the free regions sequence generation for each node along its interesting direction $r$. (c) Process of pruning infeasible paths at narrow passages. We evaluate the qualities of the free regions $\mathcal{Q}_A$, $\mathcal{Q}_B$, and their intersection $\mathcal{Q}_{A,B}$ to ensure the safe transition from \(\mathcal{Q}_A\) to \(\mathcal{Q}_B\). We search among \(a-e\) (vertices of the intersection $\mathcal{Q}_{A,B}$) to find the shortest line segment \(be\) (shown in red) that intersects the reference path from $\mathcal{Q}_A$ to $\mathcal{Q}_B$ (the line segments connecting $\boldsymbol{C}(\mathcal{Q}_{A})$, $\boldsymbol{C}(\mathcal{Q}_{A,B})$, and $\boldsymbol{C}(\mathcal{Q}_{B})$). 
}

\label{fig:extract_direction}
\end{figure}
\subsubsection{Interesting Directions Extraction}
At the replan point $\boldsymbol{p}_{n_i}^r$ of $n_i$, we first extract $k$ feature points (shown as the red dots at the obstacle corners in Fig.~\ref{fig:extract_direction}(a)) directly from the point cloud data $\boldsymbol{\mathcal{P}}_{\boldsymbol{p}_{n_i}^r}$ based on the smoothness information utilizing the algorithm outlined in \cite{zhang2014loam}. Then, $r$ interesting directions (green lines in Fig.~\ref{fig:extract_direction}(a)) are identified by evaluating the relationships between adjacent feature points based on three specific rules. The principle behind these rules is to extract directions that are worth exploring and non-redundant in the surrounding area to represent the topology of collision-free space, which shares similarity with conventional frontier points detection methods used widely in exploration area \cite{Batinovic-RAL-2021}. The process is depicted in Fig.~\ref{fig:extract_direction}(a). 
Denote $\theta_{i,j}$ the angle between two adjacent feature points $i$ and $j$, 
starting from one feature point, we sequentially evaluate the obstacle information between two consecutive feature points clockwise and generate all interesting directions with the following rules. 
\begin{itemize}
     \item  \textbf{Rule 1}:
      If two adjacent feature points are blocked by a smooth-surfaced obstacle, such as  \normalsize{\textcircled{\footnotesize{1}}} $\rightarrow$ \normalsize{\textcircled{\footnotesize{2}}}, \normalsize{\textcircled{\footnotesize{3}}} $\rightarrow$ \normalsize{\textcircled{\footnotesize{4}}}, and \normalsize{\textcircled{\footnotesize{5}}} $\rightarrow$ \normalsize{\textcircled{\footnotesize{6}}}, we do not generate interesting directions between them.
    \item  \textbf{Rule 2}: When a sudden change or jump in the depth of obstacle points is identified between two feature points, such as \normalsize{\textcircled{\footnotesize{2}}} $\rightarrow$ \normalsize{\textcircled{\footnotesize{3}}}, we consider that there may be a potential navigable path between these two points. Therefore, we generate an interesting direction 
    $a$.
    \item  \textbf{Rule 3}: In situations where no obstacles are presented between two consecutive feature points, such as \normalsize{\textcircled{\footnotesize{4}}} $\rightarrow$ \normalsize{\textcircled{\footnotesize{5}}} and \normalsize{\textcircled{\footnotesize{6}}} $\rightarrow$ \normalsize{\textcircled{\footnotesize{1}}}, we further evaluate $\theta_{i,j}$. If this angle exceeds the set threshold $\theta_0$ (typically taking $\pi/2$), we consider it a large unknown area and extend branches on both sides since these two paths could potentially lead to different patterns in an unknown environment. Otherwise, only one direction in the middle of the crack will be generated, considering that one exploration direction is sufficient in this case.

\end{itemize}
Note that all interesting directions are offset by a bias $\alpha$ towards the free space to avoid obstruction by obstacles, which could lead to poor quality of subsequently extracted regions along these lines.

\subsubsection{Dynamic Tree Update} With the extracted interesting directions at node $i$, denoted as 
$\boldsymbol{\mathcal{R}}_{n_i} = \left\{\boldsymbol{r}_1, \boldsymbol{r}_2,...,\boldsymbol{r}_r\right\}$, we grow the tree $\mathcal{T}$ from $n_i$ to extend $r$ child nodes corresponding to each interesting direction:
$$
 \boldsymbol{\mathcal{S}}_{n_i} = \left\{n_{i,1}, n_{i,2},...,n_{i,r}\right\}.
$$
Here, to represent the tree structure, we extend the notation of a node by indicating its parent node in the subscript, e.g., $n_{i,j}$ is the $j$th child node extending from node $i$. For the $j$th node in $\boldsymbol{\mathcal{S}}_{n_i}$, we sequentially extract three overlapping polytopic free regions along the $\boldsymbol{r}_j$ using the decomposition algorithm in \cite{7839930} and store them in $\boldsymbol{\mathcal{Q}}_{n_{i,j}}=\left\{\mathcal{Q}_A,\mathcal{Q}_B,\mathcal{Q}_C\right\}$, as visualized in Fig. \ref{fig:extract_direction}(b). The replan point of $n_{i,j}$ is set as the geometric center of the middle free region at $\boldsymbol{C}(\mathcal{Q}_{B})$. 
To achieve safe and effective navigation in unknown and cluttered environments with narrow passages, it is crucial to evaluate the quality of the free regions for the robot with specific geometries to traverse, while avoiding re-exploration of previously visited areas or dead-end situations within our mapless framework.

In this sense, we introduce two additional steps to prune nodes from
$\boldsymbol{\mathcal{S}}_{n_i}$ that lead to infeasible trajectories or previously visited areas and dead ends.  For each $n_{i,j}\in \boldsymbol{\mathcal{S}}_{n_i}$, we first filter out routes that are impassable for our robot, eliminating paths that would inevitably fail if passed to the backend optimizer. This step reduces the unnecessary computational load on the optimizer.
Specifically, we calculate the volume of each free region in $\boldsymbol{\mathcal{Q}}_{n_{i,j}}$ with their intersections, and eliminate the node if these volumes are smaller than the robot's volume. 
Secondly, we observe that even if the intersection of free regions is large enough, the robot may still not be able to smoothly transition from one region to another considering its specific geometry. To address this issue, we define the intersection of $\mathcal{Q}_A$ and $\mathcal{Q}_B$ as $\mathcal{Q}_{A,B}$ and extract all the vertices in $\mathcal{Q}_{A,B}$, as indicated in Fig. \ref{fig:extract_direction}(c). From these vertices, we calculate the shortest line segment that separates $\mathcal{Q}_A\cup\mathcal{Q}_B$ into two into distinct regions and compare it to the robot's minimum cross-sectional length. This cross-section is considered the necessary path for the robot to traverse from $\mathcal{Q}_A$ to $\mathcal{Q}_B$. 
If the robot cannot pass through this segment with any posture, we deem the path non-traversable and eliminate it.

Besides, to achieve effective and efficient navigation in unknown environments with limited sensor information, we keep tracking two additional sets, $\boldsymbol{\mathcal{V}}$ and $\boldsymbol{\mathcal{D}}$, to record free regions that represent previously visited and dead-end area. After selecting the next intermediate goal $n_{i,j}$, the free region at $p^{r}_{n_{i,j}}$ will be added in the set $\boldsymbol{\mathcal{V}}$, and the node will be labeled as \texttt{visited}. 
For each $n_{i,j}\in \boldsymbol{\mathcal{S}}_{n_i}$, if the replan point $\boldsymbol{p}^{r}_{n_{i,j}}$ is within $\boldsymbol{\mathcal{V}}$ or $\boldsymbol{\mathcal{D}}$, the respective node will be excluded. Also, we calculate the angle between the interesting direction and the robot’s forward direction. If this angle exceeds a predefined threshold, the direction is filtered out to prevent redundant backward exploration.

Following the two pruning steps, the truncated set of $\boldsymbol{\mathcal{S}}_{n_i}$ is:
$$
    \hat{\boldsymbol{\mathcal{S}}}_{n_i} = \left\{n_{i,1}, n_{i,2},...,n_{i,l}\right\},
$$
with $l\leq r$. If $\hat{\boldsymbol{\mathcal{S}}}_{n_i}=\emptyset$, we consider $n_i$ as non-extendable and mark it as $\texttt{dead}$. In this case, the free region at $\boldsymbol{p}^{r}_{n_i}$ will be stored in the set $\boldsymbol{\mathcal{D}}$.
\normalem
\begin{algorithm}[t]\small
\caption{Dynamic Tree Update}
\label{alg:cfto}

\KwIn{Tree $\mathcal{T}$, current node $n_i$, point cloud data $\boldsymbol{\mathcal{P}}_{\boldsymbol{p}_{n_i}^r}$, dead-end regions $\boldsymbol{\mathcal{D}}$, visited regions $\boldsymbol{\mathcal{V}}$}
\KwOut{Updated tree $\mathcal{T}$, updated dead-end regions $\boldsymbol{\mathcal{D}}$}
\DontPrintSemicolon
$\boldsymbol{\mathcal{F}}_{\boldsymbol{p}_{n_i}^r} \gets$ {Extract $k$ feature points base on $\boldsymbol{\mathcal{P}}_{\boldsymbol{p}_{n_i}^r}$} ;\;
$\boldsymbol{\mathcal{R}}_{n_i} \gets$ {Identify $r$ interesting directions from $\boldsymbol{\mathcal{F}}_{\boldsymbol{p}_{n_i}^r}$};\;
$\boldsymbol{\mathcal{Q}}_{n_{i,r}}, \boldsymbol{p}_{n_i}^r \gets$ {Generate sequences of polytopic free regions along each $\bold{r}_r \in \boldsymbol{\mathcal{R}}_{n_i}$ and associate the replan point};\;
$\boldsymbol{\mathcal{S}}_{n_i} \gets$ {Add each child node of $n_i$};\;
$\hat{\boldsymbol{\mathcal{S}}}_{n_i} \gets$ Prune non-traversable nodes using $\boldsymbol{\mathcal{Q}}_{n_{i,r}}$ and \texttt{visited/dead} nodes with set $\boldsymbol{\mathcal{D}}$ and $\boldsymbol{\mathcal{V}}$;\;
\If{$\hat{\boldsymbol{\mathcal{S}}}_{n_i} = \emptyset$}{
    Mark node $n_i$ as \texttt{dead};\;
    Update dead-end regions $\boldsymbol{\mathcal{D}}$;\;
}
\end{algorithm}
\ULforem
For each remaining child node in $\hat{\boldsymbol{\mathcal{S}}}_{n_i}$, we establish an edge $e_{i,j}$ by sequentially connecting the geometric centers of the regions in $\boldsymbol{\mathcal{Q}}_{n_{i,j}}$ through their intersections, as illustrated in Fig. \ref{fig:extract_direction}(c). These resulting line segments serve as the reference path for traversing between these two nodes, with the edge length $\ell_{e_{i,j}}$ defined as the cumulative length of these segments. Through this process, both geometrical and topological information of the collision-free space are incrementally encoded in 
$\mathcal{T}$. This enables robots to continuously select intermediate goals, facilitating efficient and safe navigation towards the destination. The comprehensive procedure is succinctly presented in Algorithm \ref{alg:cfto}.

\begin{figure}[t]	
	\centering
        \vspace{3pt}
	\includegraphics[trim=0.1cm 0.1cm 0 0.1cm, clip,width=1\linewidth]{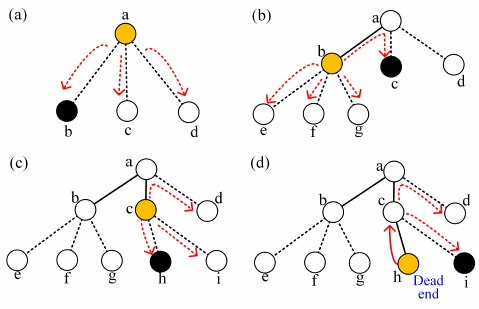}
	\setlength{\abovecaptionskip}{-0pt} 
	\caption
{
An example of intermediate goal selection during navigation. At the current node (shown in yellow), we first add all the child nodes and the second-best child node of its parent (if it exists) to a candidate intermediate goal set $\boldsymbol{\mathcal{M}}$ (the red dashed lines). We then select the node with the minimum estimated cost from $\boldsymbol{\mathcal{M}}$ as the current intermediate goal (the black node). Black dashed lines represent unexplored nodes in $\mathcal{T}$, while black solid lines represent visited paths. Notably, in unknown environments, if a dead end is encountered as in (d), the process backtracks (the red solid lines) till it finds the parent node with other feasible nodes for further exploration using the connectivity information of $\mathcal{T}$.
}
	\label{fig:node_update}
\end{figure}

\subsection{Intermediate Goal Selection}
In this section, we introduce the criteria for selecting the intermediate goal at each replanning phase. Based on the updated tree $\mathcal{T}$, we adopt a greedy search strategy by choosing the next forwarding node with the minimum cost from a candidate queue $\boldsymbol{\mathcal{M}}$. The cost of choosing a node is divided into two parts. The first part is the distance from the current replan point to the replan point of the selected node (geometric center of the middle free region along that direction), through the free regions corridor. This distance is then added to the straight-line distance from the geometric center of the furthest free region directly to the goal, serving as the esteemed cost-to-go from the selected node. In this strategy, we assume that unexplored and unseen areas are free when calculating the cost-to-go, which helps the robot to converge to the destination efficiently from the current position. Additionally, to handle common bug traps in unknown and cluttered environments, we introduce a backtracking mechanism based on the tree's structure to the next best route from the previous node to continue exploration. We will present the idea using the example process visualized in Fig. \ref{fig:node_update} for clarity.

Starting from $n_a$, as shown in Fig. \ref{fig:node_update}(a), the set of all candidate intermediate goals is defined as:
$$\boldsymbol{\mathcal{M}}_a=\left\{n_b,n_c,n_d\right\},$$
which is ordered by their corresponding cost:
\begin{equation*}
\begin{array}{cc}
    \boldsymbol{\mathcal{L}}_a &=\left\{\ell_{e_{a,b}}+\ell_{e_{b,goal}},\ell_{e_{a,c}}+\ell_{e_{c,goal}},\ell_{e_{a,d}}+\ell_{e_{d,goal}}\right\}.
\end{array}
\end{equation*}
Next, after arriving at the replan point of $n_b$, a new round of replanning is triggered, extending $n_e,n_f,n_g$ as shown in Fig. \ref{fig:node_update}(b). Instead of only searching among the child nodes of 
$n_b$, we also consider the possibility of selecting the parent node's suboptimal child node ($n_c$ in this case), since the limited sensor range may prevent us from seeing the entire obstacle layout in one frame. For instance, if a long wall blocks the subsequent path, going back for one step might offer a better cost. The candidate set at $n_b$ is thus: 
$$\boldsymbol{\mathcal{M}}_b=\left\{n_c,n_e,n_f,n_g\right\},$$
and the cost for traversing from $n_b$ to $n_c$ is:
$$
\ell_{b,c} = \ell_{e_{b,a}}+\ell_{a,c}.
$$
\normalem
\begin{algorithm}[t]\small
\caption{Intermediate Goal Selection}
\label{alg:goalselection}
\SetKwFunction{MyFunction}{CostToGo}
\SetKwProg{Fn}{Function}{:}{}
\KwIn{Tree $\mathcal{T}$, robot current position $p_r$, goal position $p_g$, current node $n_i$, dead-end regions $\boldsymbol{\mathcal{D}}$, visited regions $\boldsymbol{\mathcal{V}}$}
\KwOut{Updated tree $\mathcal{T}$, updated dead-end regions $\boldsymbol{\mathcal{D}}$, updated visited regions $\boldsymbol{\mathcal{V}}$, intermediate node $n_j$}
\DontPrintSemicolon
\If{$n_i$ is marked as \texttt{dead}}{
    $n_i \gets$ Parent node of $n_i$;\;
    \While{$n_i$ has no other feasible child node}{
        $n_i \gets$ Parent node of $n_i$;\;
    }
}
$\boldsymbol{\mathcal{M}}_i \gets$ Get candidate intermediate goals of $n_i$;\;
$\boldsymbol{\mathcal{L}}_i \gets$ Compute cost of each candidate node in $\boldsymbol{\mathcal{M}}_i$;\;
$\boldsymbol{\mathcal{M}}_i \gets$ Sort intermediate goals based on $\boldsymbol{\mathcal{L}}_i$;\;
$n_j \gets$ Select intermediate goal with minimum cost from $\boldsymbol{\mathcal{M}}_i$;\;
Mark $n_j$ as \texttt{visited};\;
Update visited regions  $\boldsymbol{\mathcal{V}}$;\;
\KwRet{$n_j$}
\end{algorithm}
\ULforem
Subsequently, as illustrated in Fig. \ref{fig:node_update}(c), the situation at $n_c$ is similar to that at $n_b$, the only difference is that although $\ell_{a,b}<\ell_{a,d}$, node $b$ has already been explored and marked as $\texttt{visited}$. The free region recorded at $n_b$ helps us avoid revisiting previously explored areas, thereby preventing redundant operations and enabling more efficient exploratory navigation. Therefore, the candidate set at $n_c$ is:
\[\boldsymbol{\mathcal{M}}_c=\left\{n_h,n_i,n_d\right\},\]
from which $n_h$ is chosen at this step.

At $n_h$, no feasible child nodes are extended by the dynamic graph updating module. the robot is considered to have entered a bug trap, exemplified by the situations in areas $A$ and $B$ of Fig. \ref{fig:maze}. To address this challenge, we propose an efficient and effective autonomous backtracking mechanism leveraging the constructed $\mathcal{T}$ to escape such dead ends. The backtracking process involves the robot iteratively retracing its steps to parent nodes until it reaches a node with unexplored feasible branches. Subsequently, a new intermediate goal is selected from the remaining candidate set. As illustrated in Fig. \ref{fig:node_update}(d), backtracking to node $n_c$ suffices for further exploration towards nodes $n_i$ and $n_d$. Node $n_i$ is chosen as the next intermediate goal due to the shorter path length:
$$\ell_{c,i}<\ell_{c,a}+\ell_{a,d}.$$ To this end, the intermediate goal selection algorithm, including the backtracking mechanism, is summarized in Algorithm \ref{alg:goalselection}.
\vspace{-2.5em}
\subsection{Collision-Free Trajectory Optimization} With the next forwarding node selected, we aim to generate a safe and effective motion plan to ensure smooth traversing from the current position to the intermediate goal.
Specifically, suppose we are navigating a robot $\mathcal{B}$ from $n_i$ to $n_j$ on $\mathcal{T}$, the following trajectory optimization is formulated:
\begin{align*}\label{eqn:to}
    \displaystyle  \operatorname*{minimize}_{(\boldsymbol{q}_{\tau},\boldsymbol{u}_{\tau})\in\re^{n}\times\re^{m}}\quad & \phi_T(\boldsymbol{q}_{T})+\sum_{\tau=0}^{T-1}J_\tau\big(\boldsymbol{q}_{\tau},\boldsymbol{u}_{\tau}\big)\notag\\ 
    \operatorname*{subject\ to}\quad \,\, & \boldsymbol{q}_{\tau+1}=f\big(\boldsymbol{q}_{\tau},\boldsymbol{u}_{\tau}\big),\notag\\ & \boldsymbol{u}_{\tau} \in [\boldsymbol{u}_{lower},\boldsymbol{u}_{upper}], \notag\\
    &\quad\quad \tau = 0,1,\ldots,T-1\yesnumber\\
    & \mathcal{W}_\mathcal{B}\left(\boldsymbol{q}_{\tau}\right)\subseteq \boldsymbol{\mathcal{Q}}_{n_{i,j}},\notag\\
    &\quad\quad \tau = 0,1,\ldots,T\notag\\
    & \boldsymbol{q}_0 = \boldsymbol{p}^r_{n_i}.\notag
\end{align*}
In this optimization problem, we seek the optimal state-control trajectory $(\boldsymbol{q},\boldsymbol{u})$ over the horizon $T$, subjecting to dynamic constraints $f$, control limit constraints, and safety constraints. Notably, the safety constraints enforce that the space occupied by the robot $\mathcal{B}$ at each $\boldsymbol{q}_{\tau}$, denoted as $\mathcal{W}_\mathcal{B}\left(\boldsymbol{q}_{\tau}\right)$, to be contained within the safe corridor from $n_i$ to $n_j$, i.e., $\boldsymbol{\mathcal{Q}}_{n_{i,j}}$, which guarantees geometry-aware collision-free maneuvers along the entire trajectory. The goal constraint is only considered in the cost function since the replan point of $n_j$ may not be safe.
To accurately model the safety constraints and solve the nonlinear and nonconvex problem effectively, we formulate a Sums-of-Squares (SOS) programming problem to determine the minimum scaling factor for the free region to encompass the robot at a specific configuration~\cite{li2024collisionfreetrajectoryoptimizationcluttered}. The value and gradient information from this scaling problem is integrated into the augmented Lagrangian iterative linear quadratic regulator (AL-iLQR) based solver ALTRO \cite{8967788}, resulting in an effective and efficient bi-level pipeline to handle the implicit geometry-aware safety constraints with rapid convergence. Detailed implementations of the trajectory optimization algorithm can be found in~\cite{li2024collisionfreetrajectoryoptimizationcluttered}.

\section{Results}
In this section, we validate the effectiveness of our proposed framework for various challenging navigation tasks through both simulations and real-world experiments. 
In simulations, the framework is implemented on an Intel i5-13400F processor. We first evaluate the overall performance of the proposed navigation framework in a maze environment with narrow passages and dead ends, arising from unknown and cluttered settings. Next, we benchmark our method against several baseline methods in a random $15\,\textup{m}\times5\,\textup{m}$  forest to further highlight the contributions and advantages of our navigation framework in generating safe and efficient navigation behavior in cluttered and narrow space without any prior knowledge of the map.
Besides
, we deploy the proposed framework on a Unitree GO1 robot, with the entire system running on an Intel NUC13 with an i7-1360P processor, to navigate the robot through an unknown and cluttered indoor environment with dynamic obstacles, showcasing its practicality and robustness in real robotic applications. Finally, we have conducted additional simulations on randomly generated long-distance, narrow forest terrains with various robot shapes navigating through them. These experiments highlight the versatility of our method in accommodating different robot geometries and navigating challenging environments. The results of these experiments are available on the project website for further reference.
The trajectory optimization problem is solved using ALTRO \cite{8967788} with the safety constraint handled implicitly as described in our previous research \cite{li2024collisionfreetrajectoryoptimizationcluttered}. During the bi-level solving iterations, the certifiable safety SOS programming problem is solved using the conic programming solver COPT \cite{ge2022cardinal}.

\subsection{Simulations}

\begin{figure}[t]	
	\centering
	\includegraphics[trim=0.1cm 0.1cm 0 0.1cm, clip,width=1.0\linewidth]{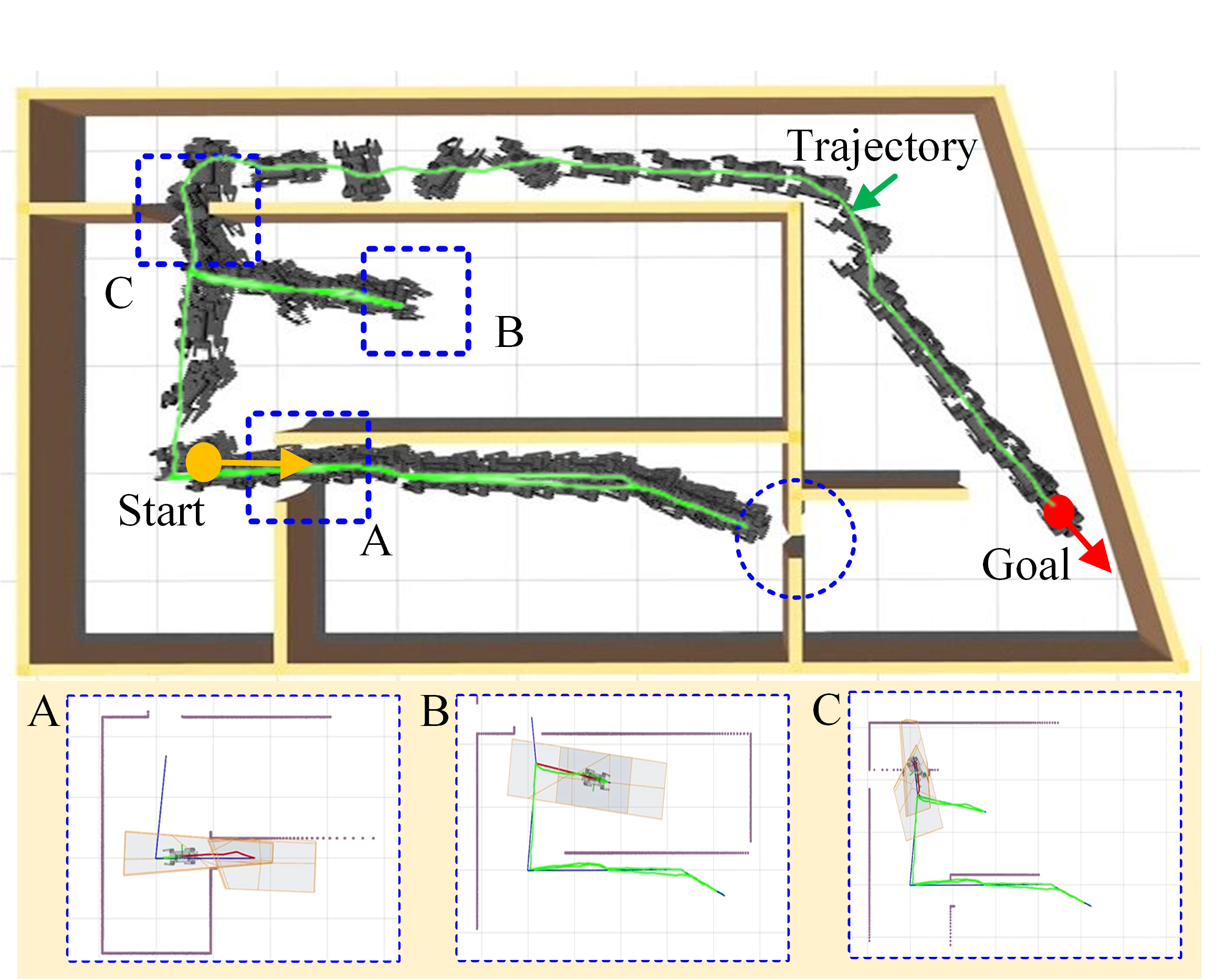}
	\setlength{\abovecaptionskip}{-10pt} 
	\caption
	{Performance of our proposed navigation framework in the maze scenario.} The overall trajectory from the start (yellow dot) to the goal (red dot) is visualized with keyframes highlighted. During navigation, the robot successfully overcomes the bug traps in frame B and the blue-circled area in region A, navigates through the narrow passages in frames A and C, and ultimately reaches the goal safely and efficiently.
	\label{fig:maze}
\end{figure}%

\begin{table}[t]
  \caption{AVERAGE COMPUTATION TIME FOR PRIMARY MODULES}
  \centering
  \scalebox{1.1}{
\begin{tblr}{
  row{1} = {c},
  row{2} = {c},
  row{3} = {c},
  row{4} = {c},
  hline{1,5} = {-}{0.08em},
  hline{2} = {-}{0.05em},
}
\textbf{Step}               & \textbf{Time} {[}ms] \\
Dynamic Tree Construction   & 0.06                \\
Intermediate Goal Selection & 0.08                \\
Trajectory Optimization     & 118                
\end{tblr}
}
\label{tab:computaion_time_maze}
\end{table}

\begin{figure*}[htp]
    	\centering
	\includegraphics[width=0.8\linewidth]{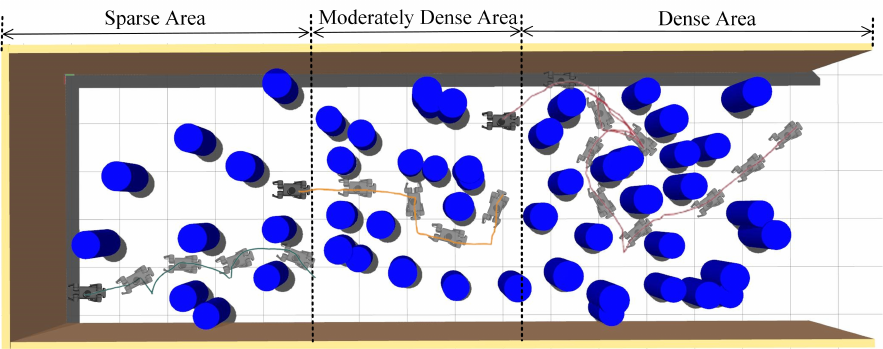}
	\caption
{
Visualization of three selected trajectories generated from our methods in the forest environment ($15\,\textup{m}\times5\,\textup{m}$). Our method efficiently and safely navigates through narrow terrains of varying obstacle densities exploiting the dynamically constructed free region tree considering specific robot geometries with no prior map.}

     \label{fig:forest}
\end{figure*}

\subsubsection{Maze}
In this subsection, we assess the performance of our proposed navigation framework in an unknown environment, as depicted in Fig. \ref{fig:maze}. Our goal is to command a $0.6\,\textup{m}\times0.4\,\textup{m}$ quadruped from the start to the goal, marked by yellow and red dots, respectively. We visualize the entire navigation process, highlighting key moments that demonstrate common challenges in cluttered and unknown environments.  Our framework relies on the available perception data and the dynamically updated free regions tree $\mathcal{T}$ to select the feasible direction with the shortest estimated cost to the goal. Consequently, the robot attempts two shorter routes to reach the goal. However, the first path is blocked by a passage narrower than the robot's width while the other path ends in a dead end. Using the dynamic tree updating rules outlined in Sec. \ref{sec:dynamic_tree}, the robot detects these dead-end situations, triggering a backtracking mechanism that allows it to retreat from these bug traps. 
Notably, due to the exploiting of the geometric relationship between the tight-fitted robot and free spaces in our navigation framework, the robot flexibly adapts its posture to navigate through narrow passages, as shown in frames A and C of Fig. \ref{fig:maze}.
The computation times of the primary modules during navigation are recorded in Table \ref{tab:computaion_time_maze}, and the entire navigation system operates in real-time at around 10 Hz.

\begin{table}[tp]
  \centering
  \caption{Comparison results of our method with RRTX \cite{rrtx} and Faster \cite{faster} in the forest environment}
    \begin{tabular}{ccccc}
    \toprule
          &       & \textbf{RRTX} & \textbf{Faster} & \textbf{Ours} \\
    \midrule
    \multirow{3}[2]{*}{\textbf{Sparse Area}} & Complete Rate & 0.7   & 0.8   & \textbf{1} \\
          & Length Scale& 1.46  & 1.45  & \textbf{1.12} \\
          & Collision Free &  \textcolor{green}{\ding{51}}      &  \textcolor{green}{\ding{51}}      &  \textcolor{green}{\ding{51}}  \\
    \midrule
    \multicolumn{1}{c}{\multirow{3}[2]{*}{\textbf{ \makecell[c]{Moderately \\ Dense Area}}}} & Complete Rate & 0     & 1     & \textbf{1} \\
          & Length Scale & N/A   & 1.73  & \textbf{1.65} \\
          & Collision Free & N/A   &  \textcolor{red}{\ding{55}}      &  \textcolor{green}{\ding{51}}  \\
    \midrule
    \multirow{3}[2]{*}{\textbf{Dense Area}} & Complete Rate & 0     & 1     & \textbf{1} \\
          & Length Scale & N/A   & 1.25  & \textbf{1.1} \\
          & Collision Free & N/A   &  \textcolor{red}{\ding{55}}      &  \textcolor{green}{\ding{51}}  \\
    \bottomrule
    \end{tabular}%
  \label{tab:forest}%
\end{table}%

\subsubsection{Forest}
\label{sec:forest}

In the forest environment, we further compare our proposed framework with RRTX \cite{rrtx} and Faster \cite{faster}. As shown in Fig. \ref{fig:forest}, to demonstrate the effectiveness and robustness of our algorithm in cluttered environments, we randomly generate obstacles with three different densities throughout the forest: 0.4 obstacles/m$^2$, 0.7 obstacles/m$^2$, 1 obstacle/m$^2$. For fairness, we set the same sensor range in the implementations of RRTX and Faster. In the three areas of varying complexity, we conduct five experiments each from different start and goal points for every method, with the completion rate, average navigation path length, and safety record in Table \ref{tab:forest}. Each successful arrival at the goal is recorded as a complete, and to standardize path length, we define the length scale as the ratio of the actual path length to the straight-line distance between each start and goal configuration. 
Typical navigation paths of our method is visualized in Fig. \ref{fig:forest}.
In moderately dense and dense areas, RRTX often fails to rewire a kinodynamically feasible path and sometimes gets stuck oscillating between two routes. Faster, as a state-of-the-art navigation framework, generally succeeds in reaching the goal in all three scenarios. It maintains a local occupancy map around the robot and uses the Jump Point Search (JPS) method to continuously search for a path to the goal. It then generates a safe corridor along this path and optimizes the robot's trajectory within the corridor. When the local map is larger than the actual map, Faster relies on the map information to escape bug traps. Both RRTX and Faster simplify the robot's shape and inflate obstacles, which in dense areas can lead to discarding the shortest feasible path for the robot's shape or failing to find a feasible route, resulting in detours and unsafe situations.

Our method does not rely on maintaining a dense map and takes the specific shape of the robot into account.
Instead, it dynamically updates a free region tree to effectively identify passable or impassable narrow gaps based on the robot's shape. In unknown environments, it continuously selects the nearest feasible path, overcomes bug traps, and optimizes a safe and effective trajectory based on the robot's tight-fitting geometry, generating non-conservative and flexible obstacle avoidance maneuvers in cluttered environments.

\subsection{Real-World Experiment}
\begin{figure}[t]
	\centering
	\includegraphics[trim=1cm 1cm 1cm 1cm, clip,width=1\linewidth]{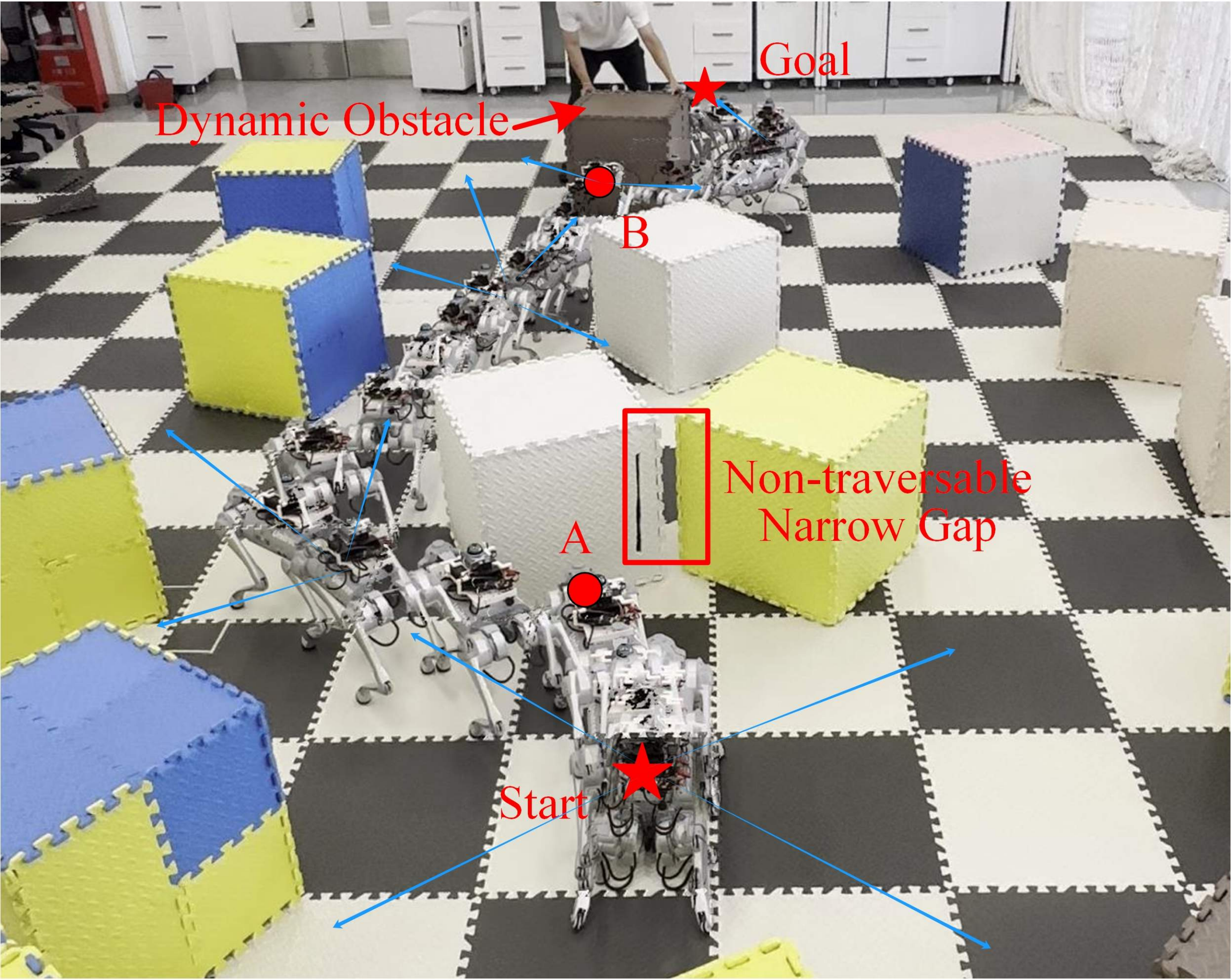}
	\setlength{\abovecaptionskip}{-0pt} 
	\caption
{Visualization of the overall trajectory in the real-world experiment. With dynamically constructed free region trees (the directions are visualized as blue arrows), the robot continuously selects suitable paths, successfully identifies impassable narrow gaps, and navigates around both static and changing environments to reach the goal.}
	\label{fig:real_exp}
\end{figure}

In this section, we deployed our framework on a unitree Go1 robot to test its performance in a cluttered indoor environment.
In the experiment, we command the robot to traverse an indoor area of $5\,\textup{m}\times6\,\textup{m}$ with several randomly placed obstacles. 
We use the onboard MID360 LiDAR to perceive the point cloud data for dynamic tree updating.
As illustrated in Fig. \ref{fig:real_exp}, the free region tree was continuously updated during the navigation process. This dynamic updating allowed the system to select suitable intermediate goals, enabling the robot to adjust its position and orientation as needed. As a result, the robot was able to successfully navigate around both static and dynamic obstacles, demonstrating its ability to adapt to changes in the environment and maintain a safe path toward its destination. 
Our system demonstrates the essential capability to mitigate real-time challenges and ensure reliable performance in complex and unpredictable settings, highlighting the effectiveness and robustness of our approach in real-world applications.
\section{CONCLUSION}
In this paper, we extend our bi-level trajectory optimization algorithm \cite{li2024collisionfreetrajectoryoptimizationcluttered} with an online replanning module for real-time, geometry-aware collision avoidance in cluttered and unknown environments. Our framework incrementally constructs a tree of free regions, efficiently representing the geometrical and topological information of the free space. During each replanning phase, exploratory paths are extended, and sequences of free regions are extracted to update the tree. The shortest feasible direction is continuously selected towards the target configuration, enabling safe and efficient navigation while adapting to environmental changes. Extensive experiments demonstrate the capability of the proposed framework in handling complex obstacle layouts and unknown terrains, ensuring safe and reliable navigation for robots with specific geometries.
\bibliographystyle{IEEEtran}
\normalem
\bibliography{ref}

\begin{thebibliography}{10}
\providecommand{\url}[1]{#1}
\csname url@samestyle\endcsname
\providecommand{\newblock}{\relax}
\providecommand{\bibinfo}[2]{#2}
\providecommand{\BIBentrySTDinterwordspacing}{\spaceskip=0pt\relax}
\providecommand{\BIBentryALTinterwordstretchfactor}{4}
\providecommand{\BIBentryALTinterwordspacing}{\spaceskip=\fontdimen2\font plus
\BIBentryALTinterwordstretchfactor\fontdimen3\font minus \fontdimen4\font\relax}
\providecommand{\BIBforeignlanguage}[2]{{%
\expandafter\ifx\csname l@#1\endcsname\relax
\typeout{** WARNING: IEEEtran.bst: No hyphenation pattern has been}%
\typeout{** loaded for the language `#1'. Using the pattern for}%
\typeout{** the default language instead.}%
\else
\language=\csname l@#1\endcsname
\fi
#2}}
\providecommand{\BIBdecl}{\relax}
\BIBdecl

\bibitem{8758904}
B.~Zhou, F.~Gao, L.~Wang, C.~Liu, and S.~Shen, ``Robust and efficient quadrotor trajectory generation for fast autonomous flight,'' \emph{IEEE Robotics and Automation Letters}, vol.~4, no.~4, pp. 3529--3536, 2019.

\bibitem{faster}
J.~Tordesillas, B.~T. Lopez, M.~Everett, and J.~P. How, ``Faster: Fast and safe trajectory planner for navigation in unknown environments,'' \emph{IEEE Transactions on Robotics}, vol.~38, no.~2, pp. 922--938, 2022.

\bibitem{10363679}
L.~Huber, J.-J. Slotine, and A.~Billard, ``Avoidance of concave obstacles through rotation of nonlinear dynamics,'' \emph{IEEE Transactions on Robotics}, vol.~40, pp. 1983--2002, 2024.

\bibitem{10607111}
Z.~Tian, Z.~Liu, X.~Zhou, and W.~Shi, ``Unguided self-exploration in narrow spaces with safety region enhanced reinforcement learning for ackermann-steering robots,'' in \emph{2024 IEEE International Conference on Mobility, Operations, Services and Technologies (MOST)}, 2024, pp. 260--268.

\bibitem{li2024collisionfreetrajectoryoptimizationcluttered}
Y.~Li, C.~Zheng, K.~Chen, Y.~Xie, X.~Tang, M.~Y. Wang, and J.~Ma, ``Collision-free trajectory optimization in cluttered environments using sums-of-squares programming,'' \emph{IEEE Robotics and Automation Letters}, 2024.

\bibitem{rrtstar}
S.~Karaman and E.~Frazzoli, ``Sampling-based algorithms for optimal motion planning,'' \emph{The International Journal of Robotics Research}, vol.~30, no.~7, pp. 846--894, 2011.

\bibitem{rrt_smart}
F.~Islam, J.~Nasir, U.~Malik, Y.~Ayaz, and O.~Hasan, ``{RRT*-Smart}: Rapid convergence implementation of {RRT\*} towards optimal solution,'' in \emph{IEEE International Conference on Mechatronics and Automation}, 2012, pp. 1651--1656.

\bibitem{local_rrt}
I.~Becerra, H.~Yervilla-Herrera, E.~Antonio, and R.~Murrieta-Cid, ``On the local planners in the {RRT*} for dynamical systems and their reusability for compound cost functionals,'' \emph{IEEE Transactions on Robotics}, vol.~38, no.~2, pp. 887--905, 2022.

\bibitem{rrtx}
M.~Otte and E.~Frazzoli, ``{RRTX}: Asymptotically optimal single-query sampling-based motion planning with quick replanning,'' \emph{The International Journal of Robotics Research}, vol.~35, no.~7, pp. 797--822, 2016.

\bibitem{9765821}
Z.~Wang, X.~Zhou, C.~Xu, and F.~Gao, ``Geometrically constrained trajectory optimization for multicopters,'' \emph{IEEE Transactions on Robotics}, vol.~38, no.~5, pp. 3259--3278, 2022.

\bibitem{9490372}
J.~Tordesillas and J.~P. How, ``{MADER}: Trajectory planner in multiagent and dynamic environments,'' \emph{IEEE Transactions on Robotics}, vol.~38, no.~1, pp. 463--476, 2022.

\bibitem{9353198}
X.~Xiao, B.~Liu, G.~Warnell, and P.~Stone, ``Toward agile maneuvers in highly constrained spaces: Learning from hallucination,'' \emph{IEEE Robotics and Automation Letters}, vol.~6, no.~2, pp. 1503--1510, 2021.

\bibitem{10417140}
Y.~Li, X.~Tang, K.~Chen, C.~Zheng, H.~Liu, and J.~Ma, ``Geometry-aware safety-critical local reactive controller for robot navigation in unknown and cluttered environments,'' \emph{IEEE Robotics and Automation Letters}, vol.~9, no.~4, pp. 3419--3426, 2024.

\bibitem{9560773}
C.~Toumieh and A.~Lambert, ``High-speed planning in unknown environments for multirotors considering drag,'' in \emph{2021 IEEE International Conference on Robotics and Automation (ICRA)}, 2021, pp. 7844--7850.

\bibitem{9718137}
J.~Park, D.~Kim, G.~C. Kim, D.~Oh, and H.~J. Kim, ``Online distributed trajectory planning for quadrotor swarm with feasibility guarantee using linear safe corridor,'' \emph{IEEE Robotics and Automation Letters}, vol.~7, no.~2, pp. 4869--4876, 2022.

\bibitem{7839930}
S.~Liu, M.~Watterson, K.~Mohta, K.~Sun, S.~Bhattacharya, C.~J. Taylor, and V.~Kumar, ``Planning dynamically feasible trajectories for quadrotors using safe flight corridors in {3-D} complex environments,'' \emph{IEEE Robotics and Automation Letters}, vol.~2, no.~3, pp. 1688--1695, 2017.

\bibitem{toumieh2022voxel}
C.~Toumieh and A.~Lambert, ``Voxel-grid based convex decomposition of 3d space for safe corridor generation,'' \emph{Journal of Intelligent \& Robotic Systems}, vol. 105, no.~4, p.~87, 2022.

\bibitem{9999335}
L.~Huber, J.-J. Slotine, and A.~Billard, ``Fast obstacle avoidance based on real-time sensing,'' \emph{IEEE Robotics and Automation Letters}, vol.~8, no.~3, pp. 1375--1382, 2023.

\bibitem{10599811}
C.~Toumieh and D.~Floreano, ``High-speed motion planning for aerial swarms in unknown and cluttered environments,'' \emph{IEEE Transactions on Robotics}, vol.~40, pp. 3642--3656, 2024.

\bibitem{8276241}
H.~Oleynikova, Z.~Taylor, R.~Siegwart, and J.~Nieto, ``Safe local exploration for replanning in cluttered unknown environments for microaerial vehicles,'' \emph{IEEE Robotics and Automation Letters}, vol.~3, no.~3, pp. 1474--1481, 2018.

\bibitem{9561460}
J.~Ji, Z.~Wang, Y.~Wang, C.~Xu, and F.~Gao, ``Mapless-planner: A robust and fast planning framework for aggressive autonomous flight without map fusion,'' in \emph{2021 IEEE International Conference on Robotics and Automation (ICRA)}, 2021, pp. 6315--6321.

\bibitem{science-gcs}
T.~Marcucci, M.~Petersen, D.~von Wrangel, and R.~Tedrake, ``Motion planning around obstacles with convex optimization,'' \emph{Science Robotics}, vol.~8, no.~84, p. eadf7843, 2023.

\bibitem{zhang2014loam}
J.~Zhang, S.~Singh \emph{et~al.}, ``{LOAM}: Lidar odometry and mapping in real-time.'' in \emph{Robotics: Science and Systems}, vol.~2, no.~9, 2014, pp. 1--9.

\bibitem{Batinovic-RAL-2021}
A.~Batinovic, T.~Petrovic, A.~Ivanovic, F.~Petric, and S.~Bogdan, ``A multi-resolution frontier-based planner for autonomous 3d exploration,'' \emph{IEEE Robotics and Automation Letters}, vol.~6, no.~3, pp. 4528--4535, 2021.

\bibitem{8967788}
T.~A. Howell, B.~E. Jackson, and Z.~Manchester, ``{ALTRO}: A fast solver for constrained trajectory optimization,'' in \emph{2019 IEEE/RSJ International Conference on Intelligent Robots and Systems}, 2019, pp. 7674--7679.

\bibitem{ge2022cardinal}
D.~Ge, Q.~Huangfu, Z.~Wang, J.~Wu, and Y.~Ye, ``Cardinal {O}ptimizer {(COPT)} user guide,'' https://guide.coap.online/copt/en-doc, 2023.

\end{thebibliography}
\end{document}